\documentclass[conference]{IEEEtran}
\IEEEoverridecommandlockouts
\usepackage{cite}
\usepackage{comment}
\usepackage{amsmath,amssymb,amsfonts}
\usepackage{algorithmic}
\usepackage{graphicx}
\usepackage{textcomp}
\usepackage{xcolor}
\usepackage{hyperref}
\usepackage{booktabs}
\def\BibTeX{{\rm B\kern-.05em{\sc i\kern-.025em b}\kern-.08em
    T\kern-.1667em\lower.7ex\hbox{E}\kern-.125emX}}

\begin{document}

\title{
    \vspace{-1em}
    {\fontsize{8pt}{10pt}\selectfont Accepted to appear at the 5\textsuperscript{th} International Conference on Robotics, Automation, and Artificial Intelligence (RAAI 2025), Singapore.}
     \\ \vspace{0.1em}
    Machine Vision-Based Surgical Lighting System: Design and Implementation
}

\author{ 
\IEEEauthorblockN{\hspace{-1.8cm}1\textsuperscript{st} Amir Gharghabi}
\IEEEauthorblockA{\hspace{-1.8cm}\textit{Faculty of Electrical Engineering} \\\hspace{-1.8cm}
\textit{Iran University of Science and Technology}\\\hspace{-1.8cm}
Tehran, Iran \\\hspace{-1.8cm}
amir.gharghabi@gmail.com}
\and 
\IEEEauthorblockN{\hspace{-1cm}2\textsuperscript{nd} Mahdi Hakiminezhad}
\IEEEauthorblockA{\hspace{-1cm}\textit{General Surgery Department} \\\hspace{-1cm}
\textit{WWL NHS Trust}\\\hspace{-1cm}
Wigan, United Kingdom \\
mahdi.hakiminezhad@gmail.com}
\and
\IEEEauthorblockN{3\textsuperscript{rd} Maryam Shafaei}
\IEEEauthorblockA{\textit{Neurology Department} \\
\textit{Imam Khomeini Hospital}\\
Tehran, Iran \\
dmr.shafaei@gmail.com}
\and
\IEEEauthorblockN{\hspace{0.8cm}4\textsuperscript{th} Shaghayegh Gharghabi}
\IEEEauthorblockA{\hspace{0.8cm}California, United States\\
\hspace{1.3cm}sh.gharghabi@gmail.com}
}

\maketitle
\begin{abstract}
	Effortless and ergonomically designed surgical lighting is critical for precision and safety during procedures. However, traditional systems often rely on manual adjustments, leading to surgeon fatigue, neck strain, and inconsistent illumination due to drift and shadowing. To address these challenges, we propose a novel surgical lighting system that leverages the YOLOv11 object detection algorithm to identify a blue marker placed above the target surgical site. A high-power LED light source is then directed to the identified location using two servomotors equipped with tilt-pan brackets. The YOLO model achieves 96.7\% mAP@50 on the validation set consisting of annotated images simulating surgical scenes with the blue spherical marker. By automating the lighting process, this machine vision-based solution reduces physical strain on surgeons, improves consistency in illumination, and supports improved surgical outcomes.

\end{abstract}


\section{Introduction}
\label{sec_intro}
Surgical lighting system (SLS) is an essential component of any operating room, enabling surgeons to carry out procedures with precision and clarity. The quality of lighting directly impacts the surgeon’s ability to visualize the surgical site, which in turn affects the accuracy of the procedure and the safety of the patient. Inadequate or improper lighting can lead to errors, complications, and even longer surgery times. Effective surgical lighting must offer optimal illumination without creating shadows or glare, ensuring the surgeon has clear visibility throughout the procedure~\cite{hemphala2020towards}.

However, traditional surgical lighting systems often face several challenges that negatively impact their performance. One notable issue is the discomfort caused by poor lighting design, which can affect the surgeon's ability to perform precise tasks. In many operating rooms, the positioning of lighting fixtures may create uneven illumination, which contributes to visual strain and reduces the clarity of the surgical field ~\cite{knulst2011indicating}. While increasing the intensity of lighting can improve contrast and enhance visibility, more lighting is not always beneficial. In the short term, excessive lighting can cause eye fatigue and headaches due to the high level of brightness, while in the long term, it may lead to eye strain and potential damage to the retina ~\cite{curlin2020current}. This issue becomes particularly significant during long and complex procedures where maintaining visual accuracy is critical.
\begin{figure}[h!]
	\centering
	\includegraphics[width=0.9\linewidth]{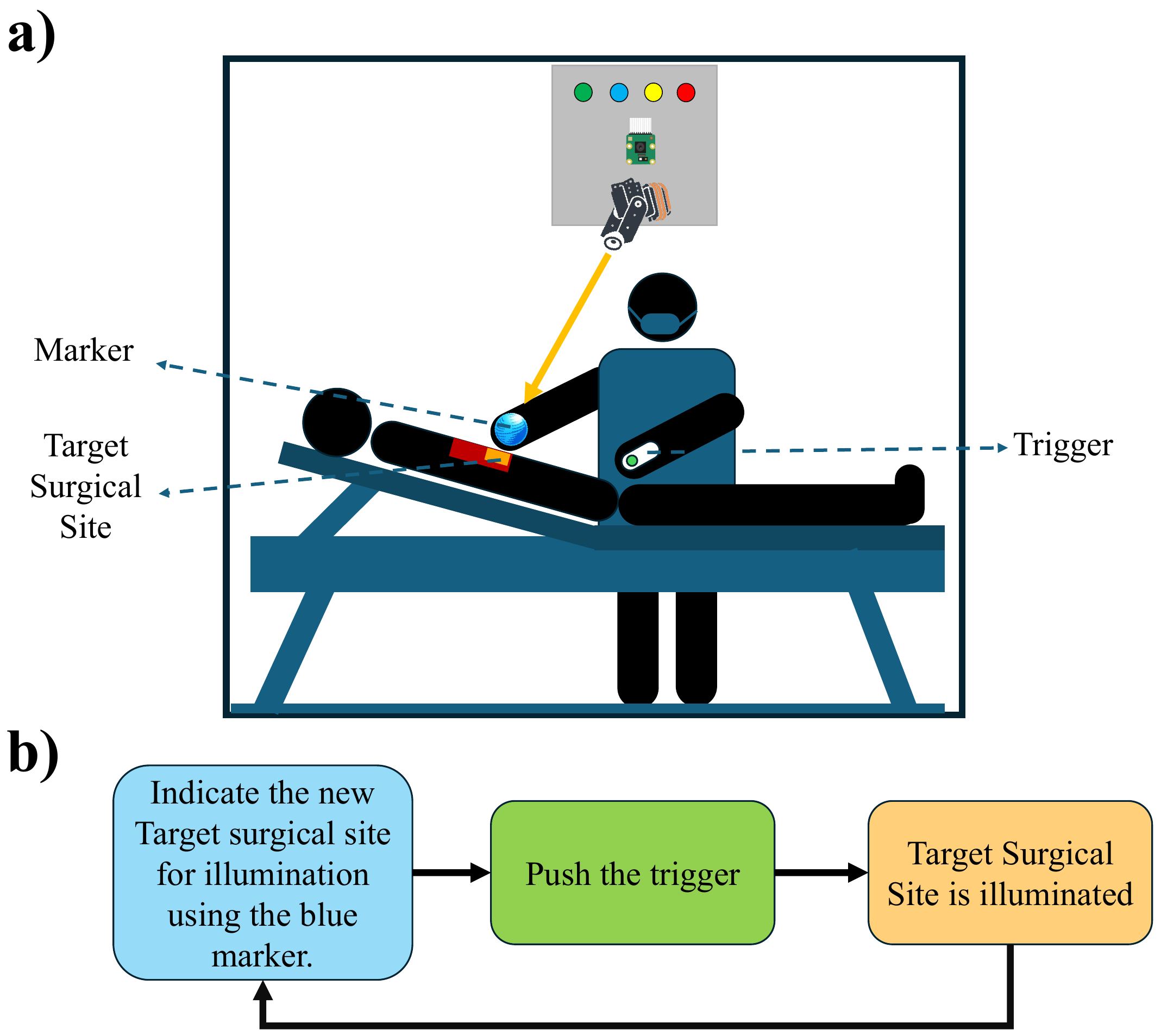}
	\caption{\centering \textbf{Overview of the proposed surgical lighting system. a)} The surgeon holds the blue marker slightly above the TSS (the orange area) and presses a trigger. The on-device camera detect and localizes the marker position, and two servomotors move the lighting package to illuminate the TSS. \textbf{b) } Sequential Workflow of the system}
	\label{fig:Intro_flowchart}
\end{figure}
In addition, conventional overhead lighting systems such as halogen lamps generate significant heat, causing discomfort for the surgical team and contributing to overheating of the operating room. This makes temperature regulation difficult, which, according to NICE Guidelines, increases the risk of intraoperative hypothermia  ~\cite{crosland2016hypothermia}. Hypothermia can cause complications such as arrhythmias, increased oxygen demand after surgery ~\cite{penjore2025effects}, and a higher risk of infection in patients ~\cite{cheadle2006risk}.

Surgeon fatigue is an increasingly recognized issue in the medical field. Long hours spent under intense lighting can cause considerable physical discomfort, particularly in the neck, back, and eyes. Continuous exposure to such conditions can lead to musculoskeletal strain and overall fatigue, which not only reduces comfort but also affects a surgeon’s focus and cognitive function during procedures~\cite{knudsen2014musculoskeletal,stucky2018surgeon}
Additionally, as Jahnavi et al. note, surgeons need to adjust overhead lamps every 7.5 minutes ~\cite{knulst2011indicating} that can contribute to distractions. This readjustment may be required due to the dynamic nature of surgery or because the lamps, being mostly mechanical tools, can experience drift errors. Furthermore, adjusting the light handles, as shown in ~\cite{cutler2020sources}, resulted in moderate field contamination, posing a potential risk to patient safety.

Current surgical lighting systems, while functional, often fail to address the full spectrum of challenges faced in the operating room. The inability to effectively eliminate shadows, combined with excessive heat generation and discomfort to the surgical team, creates an urgent need for a more efficient and comfortable lighting solution.

These limitations highlight a significant gap in the existing lighting technologies, prompting the need for an innovative approach. Traditional lighting systems have not evolved sufficiently to meet the changing needs of modern surgery, particularly as longer surgeries become more common. Therefore, the development of a new surgical lighting system that minimizes shadowing, reduces heat production, and enhances overall comfort for the surgeon is essential to improving both surgical performance and patient safety.

To achieve this goal, this paper presents a novel approach to surgical lighting through the design and implementation of an advanced system that uses the object detection algorithm (YOLO). YOLO (You Only Look Once) is a real-time object detection algorithm that processes images quickly, identifying the location of a single blue marker in a given image ~\cite{redmon2016you}. The proposed system features a small, lighting package that is dynamically directed to the target sugical site (TSS) based on the surgeon’s guidance. The surgeon holds a blue marker above the surgical site and presses a wireless button to trigger the automated lighting process. The button communicates with a Raspberry Pi using Message Queuing Telemetry Transport (MQTT), one of the most widely used protocols in Internet of Things (IoT) applications~\cite{hunkeler2008mqtt}. After detecting the blue marker by the YOLO algorithm, the system activates two servo motors to precisely direct the lighting package where it is needed. This innovative method reduces the need for manual adjustments, ensuring optimal lighting and enhancing both surgeon comfort and patient safety during long or complex procedures. A schematic illustration of the proposed method is provided in Figure~\ref{fig:Intro_flowchart} which illustrates the surgeon's interaction with the system.

The structure of this paper is as follows: Section~\ref{sec_related_work} reviews related work and recent advances in surgical lighting systems. Section~\ref{sec_methodology} details the proposed method, describing the design and implementation of our lighting system. Section~\ref{sec_results_and_discussion} presents the results and discussion, and Section~\ref{sec_limitations_future_work} outlines our limitations and future work, concluding with suggestions for further improvements.

\section{Related Work}
\label{sec_related_work}
Various approaches have been explored to enhance SLS, focusing on improving visibility, reducing manual adjustments, and increasing adaptability to different surgical environments. While traditional systems have served the medical community for years, emerging technologies are now addressing their shortcomings.

Choi (2007) developed an automated surgical illumination system using a 5-degree-of-freedom (5-DOF) robotic manipulator and ultrasonic sensors for localization to enhance lighting during surgery. The system autonomously tracked the surgeon's position and posture, adjusting the light’s position and orientation to reduce manual repositioning and minimize shadows. Their proposed system tracks the surgeon’s movements to determine their position and orientation, helping to reduce shadows on the surgical field. However, it does not aim to precisely illuminate the target surgical site (TSS) as specified by the surgeon ~\cite{choi2007automationn}.

Further advances were proposed by Teuber et al.. (2015), who introduced an autonomous surgical lamp positioning system using a depth camera. This system tracked the surgical situs and surrounding environment, adjusting the position of lamps mounted on robotic arms to ensure optimal lighting while avoiding occlusions and shadows. The optimization algorithm of the system to find the optimal position of the lamps  provided dynamic adjustments based on real-time depth data, ensuring better illumination during open surgeries ~\cite{teuber2015autonomous}.

Dietz et al. (2016) proposed a gesture-controlled surgical lighting system using a Kinect sensor, enabling sterile adjustment of illumination level, color temperature, and camera functionality for documentation. The system improved ergonomics and reduced distraction compared to conventional controls ~\cite{dietz2016contactless}.

Mühlenbrock etal. (2024) took another step forward with their modular, autonomous lighting system, designed to eliminate shadows and minimize the need for manual adjustments during surgery. Their system consists of multiple small, ceiling-mounted light modules, whose orientation and intensity are automatically controlled based on real-time data from depth sensors. This system optimizes lighting coverage and compensates for occlusions caused by surgical staff or instruments, demonstrating improved performance in illumination uniformity, particularly in open abdominal surgeries ~\cite{muhlenbrock2025novel}.

While these developments represent significant progress in surgical lighting technology, challenges remain in achieving highly precise and adaptable illumination that meets the specific needs of the surgeon during a procedure. The majority of existing methods rely on depth cameras, ultrasonic sensors, other types of proximity sensors, or manual adjustments to mitigate shadowing and ensure consistent lighting. However, depth cameras often suffer from limited accuracy in complex environments~\cite{horaud2016overview}, especially when dealing with reflective or absorptive surfaces common in operating rooms. They can also be sensitive to lighting conditions, reducing their reliability during dynamic procedures. Similarly, ultrasonic sensors face limitations in spatial resolution, are prone to interference from ambient noise, and struggle with precise localization of small targets~\cite{qiu2022review}. Notably, none of the reviewed studies used machine vision techniques for accurate localization of the surgical situs. Therefore, this paper contributes to the field by using image processing to precisely locate the surgical situs via a blue marker and employing two servo motors with tilt-pan mechanisms to illuminate the surgeon's TSS.

\section{Methodology}
\label{sec_methodology}
\subsection{Setup}

The processing unit of the system is a Raspberry Pi 4 (8GB RAM), housed in a cooling case. The Raspberry Pi Camera Module V2 captures images for detecting a blue spherical marker. The blue color was specifically chosen for the marker due to its distinctiveness in the human body. While colors like red, yellow, and green are commonly present in human tissues, blue is not naturally found \cite{pina2017human}. This absence of blue color in the human body helps ensure that the object detection algorithm performs more reliably. By selecting a color that does not interfere with biological colors, we increase the ability of the algorithm to accurately detect the marker in various lighting conditions and environments.
Although the marker is currently spherical, the shape can be altered to a more specific form if necessary to further optimize detection or to fit different application requirements.
A NodeMCU is used for wirelessly triggering the process, sending signals to the Raspberry Pi via MQTT. It is powered by a 9V battery.

The lighting system, consisting of a power LED and a focusing lens, is mounted on two MG996R servo motors equipped with tilt-pan brackets, allowing planar movement to accurately direct light onto the TSS. The power LED is cooled using a heat sink and is powered by a converter that converts the voltage from 220V AC to 5V DC. The servo motors are controlled using the raspberry pi through a PCA 9685 module, which converts 
inter-integrated circuit (I2C) communication signals to pulse width modulation (PWM) control signals, and are connected to the body of the device using a 3D-printed socket.  This setup allows the motors to adjust the lighting system and direct it to the TSS.

\begin{figure}[h]
	\centering
	\includegraphics[width=0.9\linewidth]{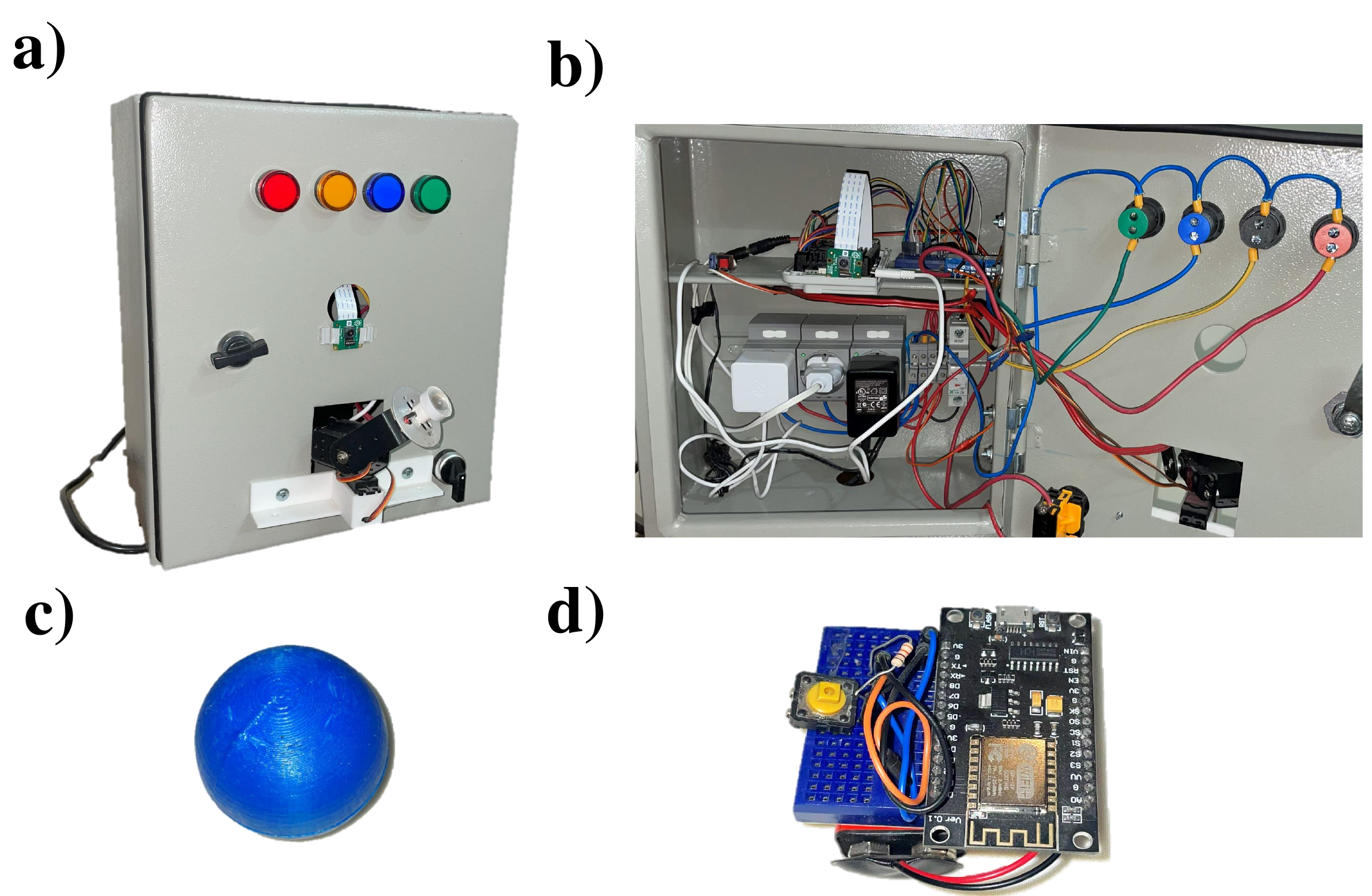}
	\caption{\centering \textbf{a)} Outer view of the device, showing the camera, servomotors carrying the lighting package, and the 4 signal lights, along with the door lock and on/off switch. \textbf{b)} Inner view of the device, showcasing the internal components. \textbf{c)} The blue spherical marker with a 3 cm diameter used for detection. \textbf{d)} The trigger mechanism using a NodeMCU and a 9V battery.}
	\label{fig:Device_in_out_blueobj_trigger}
\end{figure}

The device is housed in a steel body (30 cm x 35 cm x 15 cm) for protection and designed to be mounted on the ceiling. It includes four signal lights (red, yellow, blue, and green) to indicate various stages of operation, three industrial outlets, an industrial switch for power control, and a circuit breaker for added safety of the device. A 4-channel 5V relay module is used to control power to the signal lights. To ensure the servo motors can move freely without obstruction, a 6 cm square cutout is incorporated into the body, allowing for full range of motion without collision. A 3 cm diameter hole is considered on the body to allow the camera module, powered by the Raspberry Pi, to extend out and function. Additionally, another 3 cm diameter hole is placed on the underside of the device to route the power cable into the device. The components are mostly interfaced with the Raspberry Pi using jumper wires and a breadboard for prototyping.
A few industrial terminals are also integrated to facilitate organized and secure electrical connections for power and control lines.
Figure \ref{fig:Device_in_out_blueobj_trigger}

The device is started by simply switching the industrial power switch on. The software is configured to run automatically at boot-up on the Raspberry Pi, allowing for immediate functionality without additional input. The coding for the system was developed using Python version 3.9 to handle object detection, control of the servo motors, and communication to NodeMCU via MQTT.


\subsection{Operational Workflow}
The system utilizes a NodeMCU, equipped with a push button (trigger) that allows the operator to manually trigger the process. The NodeMCU acts as an MQTT client and continuously monitors the state of the trigger. It sends the button's state ('ON' for pressed, 'OFF' for not pressed) to the Raspberry Pi, which serves as the MQTT broker. When the trigger is pressed, the NodeMCU sends the string 'ON' to the Raspberry Pi. After receiving this signal, the Raspberry Pi initiates the image capturing process, enablingr the object detection and lighting control system to operate.
To determine the location of the blue marker, the Raspberry Pi Camera Module 2 is triggered after the button is pressed and captures exactly three images. In each image, the YOLO object detection algorithm is used to detect the blue marker and determine its center point. The average of the centers detected from the three images is then calculated to estimate the location of the marker. Capturing three images adds robustness to the process, ensuring that even if the marker is not detected in one or two images, the system can still function reliably.
For training the YOLO model, a total of 212
images were collected from the 3D printed blue spherical marker (A sphere with 3 cm diameter) using raspberry pi camera module v2. 
The dataset was divided into training, testing, and validation sets, with the specific division detailed in table ~\ref{tab:dataset_distributionnnn}. Some sample images and the link to the full dataset are provided in the Supplementary Materials.
\begin{figure}[t]
	\centering
	\includegraphics[width=1\linewidth]{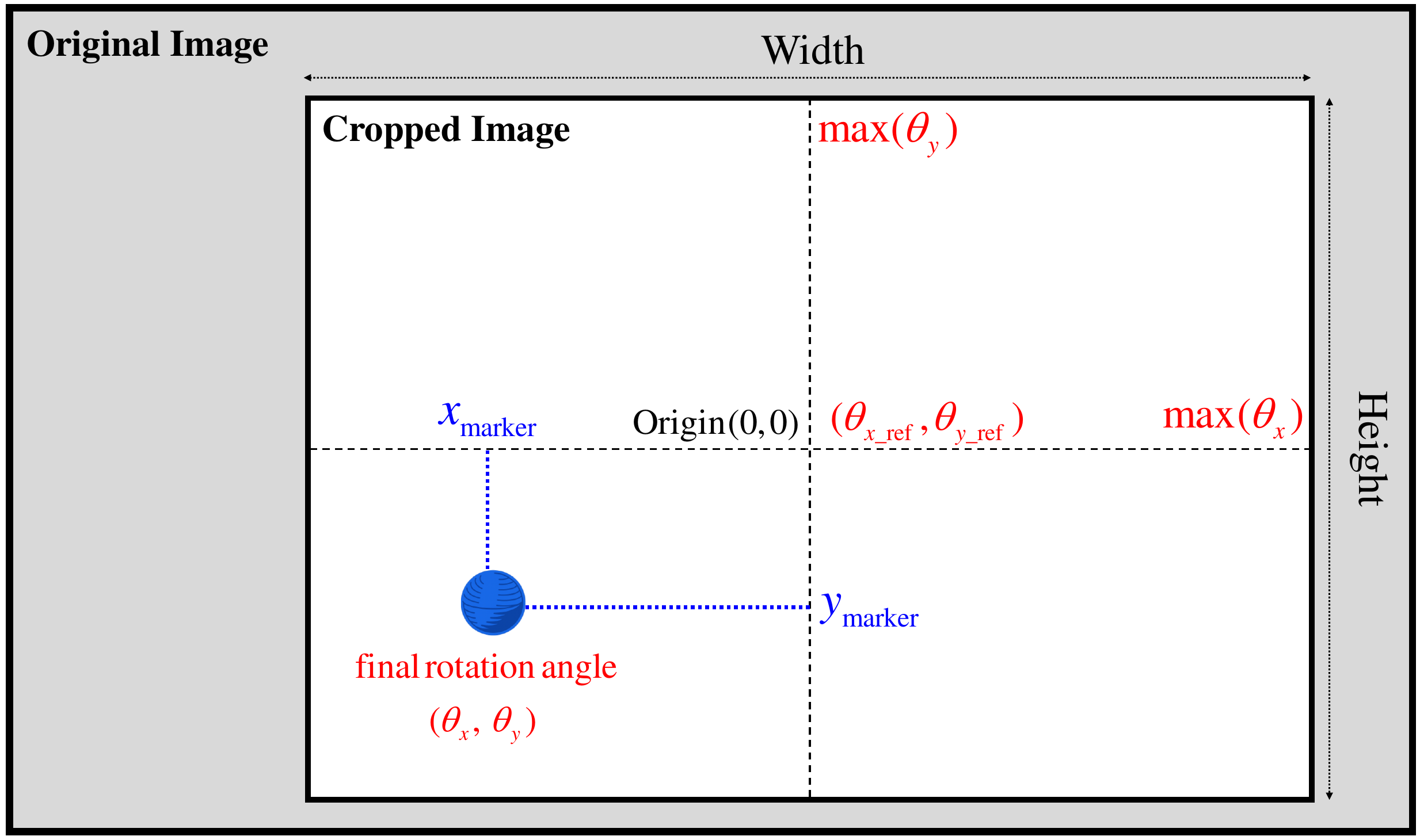}
	\caption{\centering Localization schematic depicting the camera-captured image and the cropped area representing the surgical site where the blue marker will be positioned. The schematic illustrates how the motor rotation angles are determined based on the edges of the cropped image to direct the lighting package towards the location of the blue marker.}
	\label{fig:Localization_schematic}
\end{figure}

\begin{table}
	\centering
	\caption{Distribution of images across training, testing, and validation sets}
	\begin{tabular}{l c}
		\toprule
		\textbf{Dataset} & \textbf{Number of Images} \\
		\midrule
		Train & 132 \\
		Test & 50 \\
		Validation & 30 \\
		\bottomrule
	\end{tabular}
	\label{tab:dataset_distributionnnn}
\end{table}
To localize the blue marker, the system first crops the original image using predefined parameters, specifically a starting point \((start_x, start_y)\), along with predetermined width and height values. These values ensure that the area of interest, where the blue marker is expected to be located, is captured in the cropped image. The YOLO object detection algorithm then identifies the center of the blue marker in the cropped image, producing the detected coordinates \((x_{\text{marker\_detected}}, y_{\text{marker\_detected}})\). To compute the actual coordinates of the marker, an offset is applied by subtracting half of the width and height from the detected values, resulting in the corrected marker coordinates: 
\begin{equation}
	x_{\text{marker}} = x_{\text{marker\_detected}} - \frac{\text{width}}{2},
\end{equation}

\begin{equation}
	y_{\text{marker}} = y_{\text{marker\_detected}} - \frac{\text{height}}{2}.
\end{equation}

These coordinates of the marker are then used to calculate the final angles required to guide the servomotors that carry the lighting system to illuminate the TSS. The angles are determined based on the marker's relative position to the center of the image, with adjustments made using the following equations.

\begin{equation}
	\theta_x = \theta_{x\_ref} - \left(\frac{x_{\text{marker}}}{\frac{\text{width}}{2}}\right) \cdot \theta_{x\_max},
\end{equation}

\begin{equation}
	\theta_y = \theta_{y\_ref} - \left(\frac{y_{\text{marker}}}{\frac{\text{height}}{2}}\right) \cdot \theta_{y\_max},
\end{equation}
where \(\theta_{x\_ref}\) and \(\theta_{y\_ref}\) represent the angles when the lighting is aimed at the center of the TSS, and \(\theta_{x\_max}\) and \(\theta_{y\_max}\) represent the angles when the lighting is aimed at the boundary of the TSS (Figure \ref{fig:Localization_schematic}). These angle values must be defined in calibration process before using the device.

The device is equipped with four different signal lights, each showing a specific function throughout the operation process. 
\begin{figure*}[t!]
	\centering
	\includegraphics[width=\textwidth]{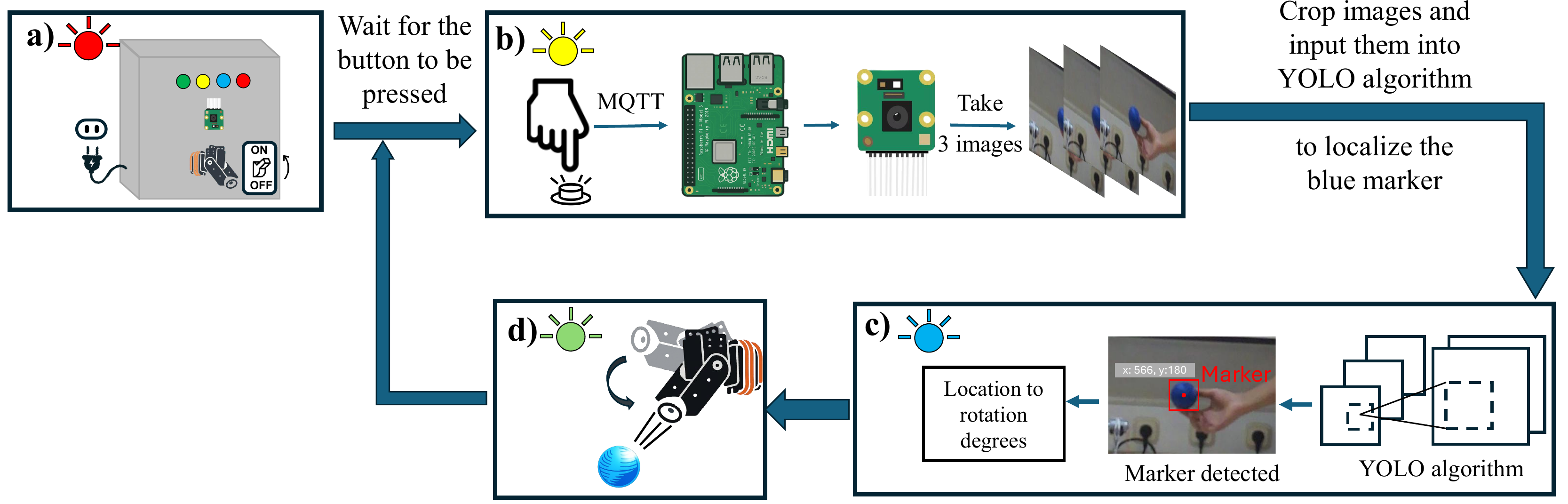}
	\caption{\centering \textbf{a)} The device is powered on and undergoes a 25-second configuration process, during which the red light is illuminated.
		\textbf{b)} Once the configuration is complete, the surgeon can hold the blue marker slightly above the TSS and presses the trigger. Upon activation of the trigger, the Raspberry Pi commands the camera to capture three images, a process that takes approximately 5 seconds. During this time, the yellow light is illuminated.
		\textbf{c)} The captured images are then processed by the YOLO algorithm to localize the blue marker. The detected location is mapped to the appropriate servo motor rotation angles, and the blue signal light turns on.
		\textbf{d)} The servo motors then adjust to the required angles, and once the motor positionin is complete, the green light is turned on. The device then waits for the next trigger press to repeat the process.}
	\label{fig:Process_flow_chart}
\end{figure*}
\begin{itemize}
	\item \textbf{Red Light}: Activated when the device is turned on, this light stays on throughout the 20-second initialization process, during which the device configuration is completed and the connection with MQTT and the NodeMCU is established. Once initialization is complete, the light turns off and remains off until the next power-up
	
	\item \textbf{Yellow Light}: When the trigger is pressed, this yellow signal light turns on to indicate that the camera is capturing images of the surgical field. During the 5-second process, the blue marker should be positioned slightly above the intended surgical site and kept steady. The yellow light remains on throughout this time, signaling that the blue marker must be held in place above the TSS.
	
	\item \textbf{Blue Light}: Turns on after the camera has captured three shots. At this point, the YOLO algorithm begins processing the images to detect the blue marker. This phase lasts for 3 seconds, with the blue light remaining on throughout the detection process. During this time, it is no longer necessary to keep the blue marker above the surgical site.
	
	\item \textbf{Green Light}: Once the YOLO detection process is completed and the servo motors have adjusted the lighting system to the TSS, the green light turns on to indicate the successful completion of the process.
\end{itemize}
The whole process is illustrated in Figure \ref{fig:Process_flow_chart}

\section{Results and Discussion}
\label{sec_results_and_discussion}

\begin{figure}
	\centering
	\includegraphics[width=0.95\linewidth]{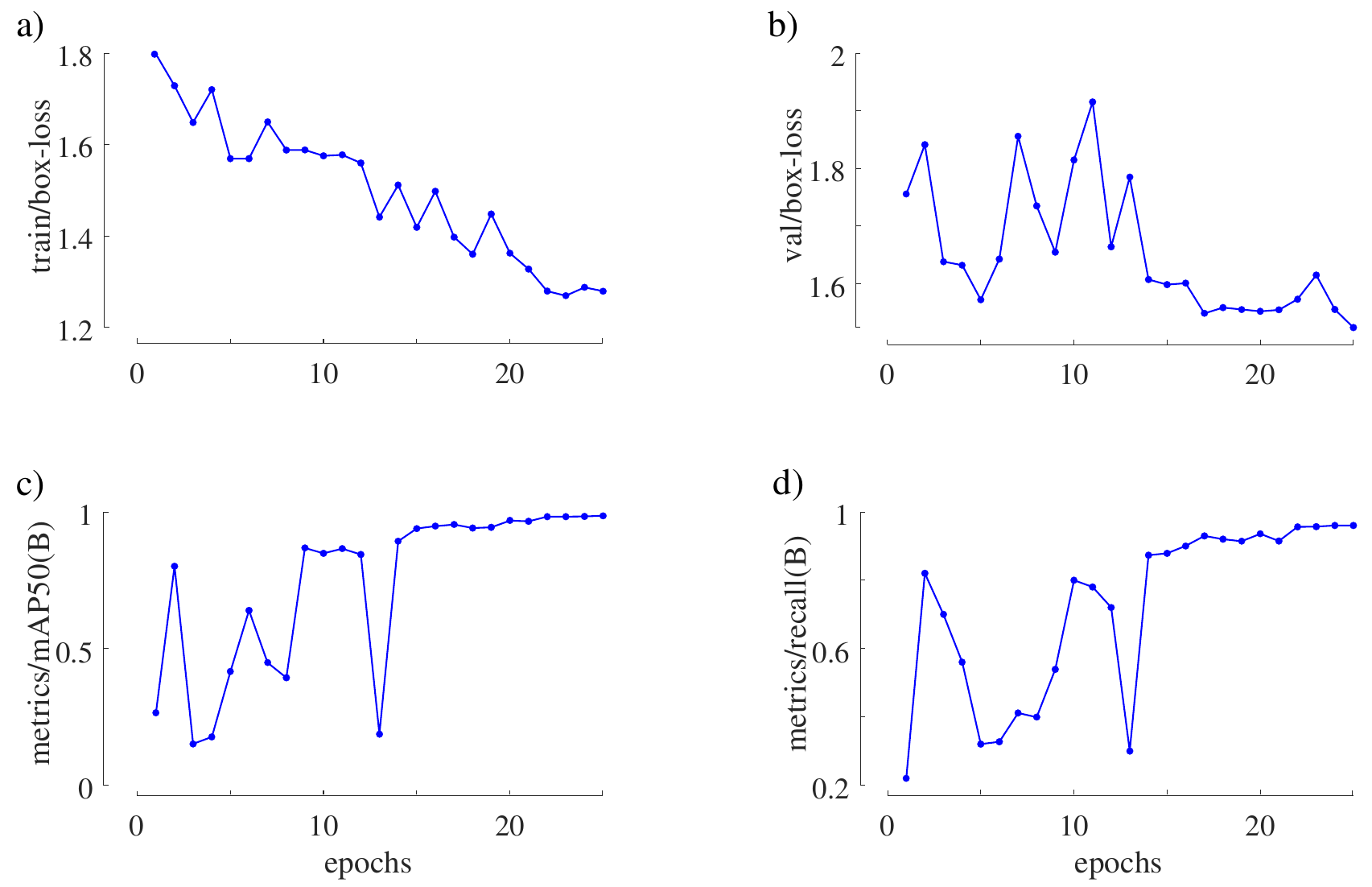}
	\caption{\centering This graph shows the performance of YOLO in detecting the blue marker over 25 epochs.\textbf{a)} Training box/loss reached 1.27 at the final epoch. \textbf{b)} Validation box/loss reached 1.52 \textbf{c)} mAP50 achieved 0.98 on test set and \textbf{d)} Recall reached 0.96}
	\label{fig:YOLO_results}
\end{figure}

The proposed method demonstrates the design and implementation of an autonomous surgical lighting system that uses machine vision to improve the efficiency and comfort of surgical procedures. The device employs the YOLOv11 object detection algorithm to detect a blue marker placed above the TSS, triggered by a wireless push button. The YOLO model achieved high performance, with 98.6\% mAP50 on the test set and 96.7\% on the validation set. The validation results indicate that the model works well on unseen data, showing no signs of overfitting. These results were anticipated, given the robust capabilities of YOLO and the simplicity of the task, as the dataset contained only one marker and was relatively small. Considering the precision on the validation set, if the blue marker is within the camera’s field of view and unobstructed, the probability of failing to detect it in all three captured images is approximately 0.003\%. However, additional data—particularly from real operating room scenes—is needed to further investigate the generalizability.
The results of training the YOLO over epochs are presented in Figure \ref{fig:YOLO_results}.

This device eliminates the need for manual lighting adjustments, allowing the surgeon to direct the light effortlessly by simply holding the blue marker and pressing the button. This automation reduces physical effort and contributes to lowering surgeon fatigue, thereby improving ergonomics during operations.

The device features a modular design and is ceiling-mounted, allowing multiple units to operate collaboratively as a multi-agent system. The lighting packages respond intelligently: if the blue marker is not visible to any unit, the corresponding lights automatically turn off. This feature enhances energy efficiency and reduces heat production. As a result, the environment around the surgical team remains cooler, decreasing discomfort. Moreover, the ability to raise the room temperature without compromising comfort helps reduce the risk of contagion by creating less favorable conditions for airborne pathogens.

In conclusion, the proposed method offers a practical and efficient alternative to conventional surgical lighting systems. By combining high-precision object detection with an intuitive control mechanism, the device improves both the surgical environment and the overall experience for medical personnel.

\section{Limitations \& Future Work}
\label{sec_limitations_future_work}
One key limitation of this study is that the system has not yet been tested in a real operating room. This is primarily due to strict hospital protocols and sterilization standards, which prevent the use of non-certified devices in sterile environments. Clinical testing would require regulatory approvals and adherence to rigorous safety and hygiene procedures, which were beyond the scope of this development phase~\cite{dhruva2024physicians}.

Future enhancements could include deploying multiple lighting units coordinated through a multi-agent system to improve adaptability and coverage. The physical size of the device can also be reduced through miniaturization. Additionally, upgrading the motors will enable more precise movement of the lighting module, and integrating a more powerful, tunable lighting source will offer greater flexibility to meet varying surgical lighting demands.

\section*{Acknowledgment}
\label{sec_acknowledgment}
We would like to acknowledge
ChatGPT for its assistance in improving grammatical
accuracy and enhancing the cohesion and coherence of
our text

\section*{Data Availability}
\label{sec_data_availability}
The full blue marker image dataset can be found using the link below.

\begin{itemize}
	\item \href{https://app.roboflow.com/surgical-lighting-system/blue-marker-dataset/browse?queryText=&pageSize=50&startingIndex=0&browseQuery=true}{Blue marker image dataset}
\end{itemize}

\label{sec_references}

\vspace{12pt}

\end{document}